\documentclass[12pt,journal,compsoc,onecolumn]{IEEEtran}
\usepackage{amsmath,graphicx,epstopdf,tabularx,multirow,colortbl,xcolor,bbding,siunitx,amssymb,stmaryrd,float,subfigure}

\usepackage[noadjust]{cite}

\newlength{\bibitemsep}
\newlength{\bibparskip}
\let\oldthebibliography\thebibliography
\renewcommand\thebibliography[1]{
  \oldthebibliography{#1}
  \setlength{\parskip}{\bibitemsep}
  \setlength{\itemsep}{\bibparskip}
}

\def\1{{\bf 1}}

\usepackage[ruled,vlined]{algorithm2e}

\title{Few-Shot Object Detection with Sparse Context Transformers}

\author{Jie Mei$^{1}$ \ \ \ \ \ \ \ \ \ \ \ \   Mingyuan Jiu$^{*1,2,3}$ \ \ \ \ \ \  \ \ \ \ \ \  Hichem Sahbi$^{*4}$  \\  $ $  \\ Xiaoheng Jiang$^{1,2,3}$ \ \ \ \ \ \  \ \ \ \ \ \  Mingliang Xu$^{1,2,3}$\thanks{This work is supported by the National Natural Science Foundation of China (No.~62272422, U22B2051).}
  $ $ \\
  $ $ \\
{\small $^{1}$ School of Computer and Artificial Intelligence, Zhengzhou University, Zhengzhou, China\\
	$^{2}$ Engineering Research Center of Intelligent Swarm Systems, Ministry of Education,\\ Zhengzhou University, Zhengzhou, China\\
	$^{3}$ National Supercomputing Center in Zhengzhou, Zhengzhou, China\\
	$^{4}$ Sorbonne University, CNRS, LIP6, F-75005, Paris, France}
        
}

\begin{document}
 \maketitle
\begin{abstract}
		Few-shot detection is a major task in pattern recognition which seeks to localize objects using models trained with few labeled data. One of the   mainstream few-shot methods is transfer learning which consists in pretraining a detection model in a source domain prior to its fine-tuning in a target domain. However, it is challenging for fine-tuned models to effectively identify new classes in the target domain, particularly when the underlying labeled training data are scarce. In this paper, we devise a novel sparse context transformer (SCT) that  effectively leverages object knowledge in the source domain, and automatically learns a sparse context from only few training images in the target domain. As a result, it  combines different relevant clues in order to enhance the discrimination power of the learned detectors and reduce  class confusion. We evaluate the proposed method on two challenging few-shot object detection benchmarks, and empirical results show that the proposed method obtains competitive performance compared to the related state-of-the-art.         
\end{abstract}

	\section{Introduction}
	\label{sec:intro}
	Although  deep learning (DL) models  have achieved remarkable performance, these outstanding models are usually  label-hungry and their training is time and memory demanding. In some   scenarios, particularly object detection, labeled data are scarce, and this makes  DL-based detection a major challenge \cite{kohler2021few, jiaxu2021comparative}.    Among existing solutions that mitigate scarcity of labeled data, transfer learning is particularly  effective and consists in pretraining detection models in the source domains --- using abundant labeled data --- prior to their fine-tuning in the target domains.    However, as labeled data  are scarce in the target domains,  few-shot object detection --- based on transfer learning  \cite{qiao2021defrcn, li2021few, Sun_Li_Cai_Yuan_Zhang_2021} --- is not yet sufficiently effective in identifying new object classes.\\
	
	\indent The aforementioned issue is  accentuated  in few-shot object detection as this task involves both {\it localization and classification}.  Localization focuses on  spatial information which is decently obtained  from pretrained models in the source domains. Therefore,   bounding box regressors (BBOX) trained in the source domain are  already reliable for initialization and fine-tuning in the target domain. Hence, detectors fine-tuned with few training samples may effectively locate new object classes. In contrast, classification often requires contextual knowledge of specific categories. In other words, source domain knowledge are insufficient to learn new category distributions in the target domain. Therefore, the underlying models should  be completely retrained for new categories. However, scarcity and limited diversity of data in the target domain lowers the accuracy of the new learned category classifiers, and thereby leads to class confusion. \\
	
	\indent In order to address these issues, we propose in this paper a novel {\it sparse context transformer} (SCT) that  leverages source domain knowledge together with few training images in the target domain. This transformer learns sparse affinity matrices between BBOXs and classification outputs by exploring the most relevant contexts for new object categories in the target domain. The proposed transformer consists of two simple yet effective submodules: (i) sparse relationship discovery, and (ii)  aggregation. In submodule (i), contextual fields are initially designed based on default prior boxes (also called anchor boxes)   \cite{Girshick_2015, Liu_Anguelov_Erhan_Szegedy_Reed_Fu_Berg_2016} and multi-scale feature maps extracted from a visual encoder. Then, relationships between each prior box and contextual fields are modeled through a novel sparse attention.  In submodule (ii), aggregation further leverages the learned relationships and integrates contextual fields into the relevant prior boxes. As shown through experiments, our proposed transformer enhances prior box representations, and mitigates  confusion in few-shot  object detection and classification. \\
	
	\noindent Considering all the aforementioned issues, our main contributions include
	\begin{itemize} 
		\item  A novel  sparse context transformer that  effectively explores useful contextual fields from a small number of labeled images. This transformer is embedded into an SSD (plug-and-play style)  detector  suitable for few-shot object detection.
		
		\item A novel attention layer that assists object detection in learning task-relevant knowledge from images by enhancing the underlying task-related feature representations.
		
		\item A comprehensive evaluation of our proposed method on the challenging configurations for few-shot detection that shows high performance.
	\end{itemize} 
	
	\section{Related Work}
	Recent years have witnessed a significant progress in few-shot object detection  using transfer learning. Among existing methods, Chen et al.~\cite{DBLP:journals/corr/abs-1803-01529} introduced a low shot transfer detector that  focuses on foreground objects in the target domain during  fine-tuning in order to learn more knowledge on the targeted categories.  Khandelwal et al.~\cite{Khandelwal_Goyal_Sigal_2021} conjectured that simple fine-tuning may lead to a decrease in the transferability of the models. Hence, they proposed a unified semi-supervised framework that combines weighted multi-modal similarity measures between base and novel classes. With this method, they achieved effective knowledge transfer and adaptation. Unlike these methods, Wang et al. \cite{Yang_Wang_Chen_Liu_Qiao_2020} proposed a context-transformer that    tackles object confusion  in few-shot detection. This  transformer relies on a set of contextual fields  from different spatial scales and aspect ratios of prior boxes, in order to explore their relationships through dot products. Based on these relationships, the contextual fields are integrated into each prior box and this improves their representation.\\      
	\indent Our work is an extension of context transformers that addresses the relatively monotonous contextual fields (constructed in the original version of these transformers) as well as their relationships with prior boxes, which cannot effectively suppress task-independent contextual fields, and further affect the model's ability to recognize novel classes. In this regard,  we consider informations from different sources and we model sparse relationships between contextual fields and each prior box to help the model selecting the most effective fields. This also mitigates confusion in few-shot object detection.

	\section{Proposed Method}
	
\begin{figure*}[htb]
		\centering
		\includegraphics[width=18.5cm]{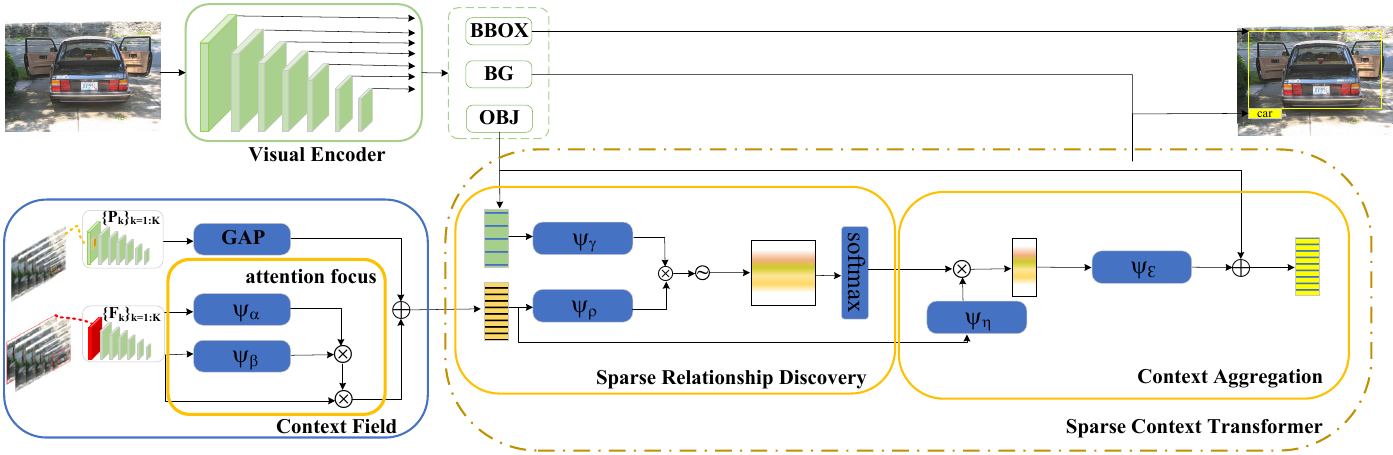}
		\caption{Few-Shot Detection with Sparse-Context-Transformer. It consists of sparse relationship discovery and context aggregation, which can effectively utilize the context fields of few-shot tasks, improve the context awareness ability of each prior box, and solve the problem of object confusion in few-shot detection. The attention focus module can effectively help us learn task-related contextual fields.}
		\label{pipeline}
	\end{figure*}

	In this section, we  introduce our novel sparse context transformer. As shown in Fig.~\ref{pipeline}, our framework relies on an SSD-style detector~\cite{Liu_Anguelov_Erhan_Szegedy_Reed_Fu_Berg_2016} used as a flexible plug-and-play backbone  that delivers rich multi-scale contextual   information.   The SSD detector  consists of $K$ (spatial-scale) heads    including bounding boxes regressor (BBOX) and object+background classifiers (OBJ+BG). To generalize few-shot learning in the target domain, we first pretrain the SSD detector with a large-scale dataset in the source domain. Then, we combine the proposed transformer module with the SSD detector for fine-tuning in the target domain (see again Fig.~\ref{pipeline}). As shown subsequently, our proposed  transformer  includes two submodules: one for sparse relationship discovery, and another one for aggregation. These submodules are respectively used to model context/classifier relationships and for context fusion. 
	
	\subsection{Sparse Relationship Discovery}
	
	Given an image $\cal I$ fed into an SSD detector, we extract for each prior box in $\cal I$ a vector of scores $\mathbf{P}_{\emph{k},m,h,w} \in \mathbb{R}^{C_s}$; being $C_s$ the number of (source) object categories, $k \in \llbracket 1,K\rrbracket$ a spatial scale,  $m$ an aspect ratio, and $(h, w)$ the prior box coordinates at the $k$-th scale. In what follows, we reshape the tensor  $(\mathbf{P}_{\emph{k},m,h,w})_{_{\emph{k},m,h,w}}$ as a matrix $\mathbf{P} \in \mathbb{R}^{D_p \times C_s}$; being  $D_p$ the total number of prior boxes in $\cal I$ across all the possible scales, aspect ratios and locations.  Scores in  $\mathbf{P}$ provide us with a rich semantic representation about object categories \cite{Tzeng_Hoffman_Darrell_Saenko_2015}; nonetheless, this representation is deprived from contextual relationships between prior boxes. Since  SSD usually involves ten thousand prior boxes per image, modeling and training all the relationships between these  boxes is clearly  intractable, overparameterized and thereby subject to overfitting, particularly in the few-shot scenario. In order to prevent these issues,  spatial pooling is first achieved so one may obtain a more compact matrix $\mathbf{Q} \in \mathbb{R}^{D_q \times C_s}$ instead of  $\mathbf{P}$, being $D_q$ ($\ll  D_p$) the reduced number of prior boxes after spatial pooling.\\
        
	\indent        Considering that prior boxes capture only object parts (and not their overall extents),   multi-scale feature maps  extracted by the SSD encoder (denoted as $\{\mathbf{F}_k\}_k$)  are also aggregated as $\textbf{M} = \textrm{Concat}(\{\mathbf{F}_{\emph{k}}\}_k) \in \mathbb{R}^{D_q \times D_f}$; being $\textrm{Concat}$ the concatenation operator and $D_f$ the resulting dimension after the application of this operator. These aggregated features $\textbf{M}$ enable to provision  with complementary visual cues at different scales, and provide us with a more comprehensive contextual information~\cite{Zhao_Shi_Qi_Wang_Jia_2017}. In the rest of this paper, the {\it  pairs} of pooled prior boxes  $\mathbf{Q}$ together with the underlying aggregated features $\textbf{M}$ are referred to as {\it contextual fields}.\\
	
	\noindent {\bf Attention Focus.} In order to learn task-related context from few training data, we design an attention focus layer that enhances the representation of contextual fields and attenuates  category confusion.  We first define the attention weight matrix ${\bf A}_{\textbf{M}}$ as
	\begin{equation}
		{\bf A}_{\textbf{M}} = \psi_{\alpha}(\textbf{M})^\top \   \psi_\beta(\textbf{M}),
		\label{attentionFocus}
	\end{equation}
	\noindent being $\psi_{\alpha}(.)$ (and $\psi_{\beta}(.)$) trained fully-connected (FC) layers that increase the expressivity of the  attention matrix	${\bf A}_{\textbf{M}}\in \mathbb{R}^{D_f \times C_s}$, and allow obtaining an enhanced representation	$\textbf{M}^{*} \in \mathbb{R}^{D_q \times C_s}$ as 
	\begin{equation}
		\textbf{M}^{*} = \textbf{M}  \  \ {\bf A}_{\textbf{M}},         
		\label{attentionFocus2}
	\end{equation}
	
	\noindent which also ensures dimension consistency of the learned representation in Eq.~\eqref{lc}. By combining the pooled prior boxes in $\mathbf{Q}$ and the underlying multi-scale feature maps  $\textbf{M}^{*}$, we obtain our contextual field representations  that capture both intrinsic (feature) and extrinsic (object-class) information, resulting into 
	\begin{equation}\label{lc}
		\textbf{C} =  \lambda \ \textbf{M}^{*} \  + \  \textbf{Q},
	\end{equation}

	\noindent here $\lambda \geq 0$ controls the impact of attention in $\textbf{M}^{*}$. Using Eq.~\eqref{lc}, we design our sparse attention mechanism in order to explore the affinity relationships between each prior box and contextual fields, and remove spurious ones (i.e., those farther away from the prior boxes) according to the sparse relationship. More precisely, we evaluate the relationship matrix between the contextual fields in $\mathbf{C}$ and prior boxes in $\mathbf{P}$, and we reset  weak  relationships to zero using soft-thresholding as  
	\begin{equation}
		\textbf{R} = \textrm{sign}(\textbf{A}),
	\end{equation}
	with 
	\begin{equation}
		\mathbf{A} =\textrm{softmax}\bigg(\frac{\psi_\gamma(\mathbf{P}) \ \psi_\rho(\textbf{C})^\top}{\sqrt{C_s}}\bigg).
	\end{equation}
	
	\noindent Here  $\textrm{softmax}$ is row-wise applied, $\psi_{\gamma}(.)$, $\psi_{\rho}(.)$ are again trainable FC layers and $\textrm{sign}(.)$ is an operator that sparsifies a given relationship matrix using a soft threshold. Each row of  $ \textbf{R} \in \mathbb{R}^{D_p \times D_q}$  measures the importance of all contextual fields w.r.t. its underlying prior box. Hence, sparse relationship discovery allows a prior box to identify its important contextual fields and  discard those that are not sufficiently important  according to various aspect ratios, locations and spatial scales.
	
	\begin{table*}
	\centering
	\resizebox{0.9\textwidth}{!}{
		\begin{tabular}{c|ccccc|ccccc|ccccc|}
			\hline\hline\noalign{\smallskip}	
			\multirow{2}{*}{Method/shot} & \multicolumn{5}{c|}{Novel Set 1}  &\multicolumn{5}{c|}{Novel Set 2} & \multicolumn{5}{c|}{Novel Set 3} \\
			\cline{2-16}
			& 1& 2& 3& 5& 10 & 1& 2& 3& 5& 10 & 1& 2& 3& 5& 10 \\
			\hline\hline\noalign{\smallskip}
			Shemelkov et al.2017\cite{Shmelkov_Schmid_Alahari_2017}
			& 23.9 & - & - & 38.8 & - 
			& 19.2 & - & - & 32.5 & - 
			& 21.4 & - & - & 31.8 & - \\
			Meta YOLO\cite{Kang_Liu_Wang_Yu_Feng_Darrell_2019} 
			& 14.8 & - & - & 33.9 & - 
			& 15.7 & - & - & 30.1 & - 
			& 19.2 & - & - & 40.6 & - \\
			Meta R-CNN \cite{Yan_Chen_Xu_Wang_Liang_Lin_2019} 
			& 19.9 & 25.5 & 35.0 & \textcolor{blue}{45.7} & \textcolor{blue}{51.5} 
			& 10.4 & 19.4 &29.6 & 34.8 & \textcolor{blue}{39.7} 
			& 14.3 & 18.2 & 27.5 & \textcolor{blue}{41.2} & \textcolor{blue}{48.1}  \\
			DCNet \cite{Hu_Bai_Li_Cui_Wang_2021} 
			& \textcolor{blue}{33.9} & \textcolor{blue}{37.4} & \textcolor{red}{43.7} & \textcolor{red}{51.1} & \textcolor{red}{59.6} 
			& 23.2 & \textcolor{blue}{24.8} & \textcolor{blue}{30.6} & \textcolor{red}{36.7} & \textcolor{red}{46.6} 
			& \textcolor{blue}{32.3} & \textcolor{red}{34.9} & \textcolor{red}{39.7} & \textcolor{red}{42.6} & \textcolor{red}{50.7}   \\
			Cos R-CNN \cite{Data_Howard-Jenkins_Murray_Prisacariu_2023} 
			& 27.9 & 33.0 & 32.1 & 36.2 & 33.6 
			& 19.4 & 12.6 & 14.4 & 19.1 & 21.9 
			& 16.9 & 21.6 & 21.6 & 27.5 & 25.5 \\
			Baseline
			& 34.2 & - & - & 44.2 & - 
			& \textcolor{blue}{26.0} & - & - & 36.3 & -
			& 29.3 & - & - & 40.8 & - \\
			SCT(ours) 
			& \textcolor{red}{37.9} & \textcolor{red}{40.8} & \textcolor{blue}{41.2} & 44.9 & 45.5
			& \textcolor{red}{32.1} & \textcolor{red}{33.1} & \textcolor{red}{32.7} & \textcolor{blue}{36.3} & 36.4
			& \textcolor{red}{33.7} & \textcolor{blue}{31.8} & \textcolor{blue}{35.4} & 39.1 & 43.2 \\
			\hline
		\end{tabular}
              }
              \vspace{0.5cm}
	\caption{Few-Shot detection performance (mAP) on PASCAL VOC dataset. We evaluate 1, 2, 3, 5, 10 shot performance over multiple runs. \textcolor{red}{red} and \textcolor{blue}{blue} indicate the best and second best results, respectively, and '-' indicates no reported results (i.e., not available).}
	\label{tab:compareVOC}
	\end{table*}
	
	\subsection{Aggregation}
	
	We consider the sparse relationship matrix $\textbf{R}$ ---  between prior boxes and contextual fields --- as a relational attention in order to derive  the representation of each prior box. We also consider a $\textrm{softmax}$ operator on each row $i$ of $\textbf{R}$ as a gating mechanism that measures how important is each contextual field w.r.t. the $i$-th prior box.  By considering the cross correlations between rows of $\textbf{R}$ and columns of $\textbf{C}$, we derive our sparse attention-based representation of prior boxes as
	\begin{equation}
		\textbf{W} = \textrm{softmax}(\textbf{R}) \ \psi_\eta(\textbf{C}),
	\end{equation}
	
	\noindent   being $\textbf{W} \in \mathbb{R}^{D_p \times C_s}$ and $\psi_\eta$ corresponds again to trainable FC layers. Now, we combine $\textbf{W}$ with the original matrix of prior boxes  $\textbf{P}$ in order to derive our final context-aware representation  ${\hat{\textbf{P}}} \in \mathbb{R}^{D_p \times C_s}$
	
	\begin{equation}
		\hat{\textbf{P}} = \textbf{P} + \psi_{\xi}(\textbf{W}).
	\end{equation}

	\noindent Here $\psi_\xi$ corresponds to other (last) trainable FC layers. Since $\hat{\textbf{P}} $ is context-aware, it enhances the discrimination power of prior boxes by attenuating confusions between object classes. By plugging the final representation $\hat{\textbf{P}}$ into a softmax classifier, we obtain our scoring function, on the $C_t$ target classes, as  
	\begin{equation}
		\hat{\textbf{Y}} = \textrm{softmax}(\hat{\textbf{P}} \  \Theta).
	\end{equation}
	\noindent  Note that the representations in $\hat{\textbf{P}}$ and the underlying parameters $\Theta \in \mathbb{R}^{C_s \times C_t}$  are shared across different aspect ratios and spatial scales, so there is no requirement to design separate classifiers at different scales. This not only reduces computational complexity but also  prevents overfitting. 	

	\section{Experiments}
	\label{sec:pagestyle}
	In this section, we introduce our experimental settings and evaluations of our method on two standard benchmarks, namely PASCAL VOC and  MS COCO.
	\subsection{Datasets and Settings}
	Following \cite{Kang_Liu_Wang_Yu_Feng_Darrell_2019}, we describe the few-shot detection datasets for comparison against the related works. All the quantitative performance correspond to the mean Average Precision (mAP) evaluation metric.\\
	
	\noindent \textbf{PASCAL VOC 2007/2012 \& MS COCO}. The train and test sets of PASCAL VOC 2007 and 2012 are split into source (seen) and target (unseen) object categories. Three different splits are considered for the unseen categories, namely    ["bird", "bus", "cow", "motorbike","sofa"], ["aeroplane", "bottle", "cow", "horse","sofa"], and ["boat", "cat", "motorbike", "sheep" , "sofa"].   	As for MS COCO~\cite{Lin_Maire_Belongie_Hays_Perona_Ramanan_Dollar_Zitnick_2014}, it includes 80 object categories where 20 of them  --- that overlap with PASCAL VOC --- are used as unseen categories. The process of constructing the few shot dataset, and seen/unseen categories in MS COCO is similar to PASCAL VOC. \\ 
	
	\noindent\textbf{Implementation Details}. We choose a recent SSD detector~\cite{Liu_Huang_Wang_2018} as a basic architecture built upon 6 heads corresponding to different spatial rescaling factors (taken in $\{1,3,5,10,19, 38\}$). The contextual fields we designed consist of two parts: in the first one, prior boxes --- corresponding to multiple scales and aspect ratios --- are max-pooled with different instances of kernel sizes+strides taken in $\{2, 3\}$. For the second one, contextual fields composed of multi-scale features are fused through four spatial scales. In these experiments, we set the hyperparameter $\lambda$ in Eq.~\eqref{lc} to 0.6, and the embedding functions in the sparse context transformer correspond to the residual FC layer whose input and output have the same number of channels.\\              
	\noindent We implement our experiments using PyTorch~\cite{Paszke_Gross_Chintala_Chanan_Yang_DeVito_Lin_Desmaison_Antiga_Lerer_2017} on two Nvidia 3090 GPUs. We pretrain the SSD detectors on the source domain following exactly the original SSD settings in~\cite{Liu_Huang_Wang_2018}, and we fine-tune these SSDs on the target domain using stochastic gradient descent with the following settings: a batch size of 64, a momentum of 0.9, an initial learning rate equal to $4\times 10^{-3} $ (decreased by 10 after 3k and 3.5k iterations), a weight decay of $5\times 10^{-4} $, and a total number of training iterations equal to 4k.
	
	\subsection{Results on PASCAL VOC}
	In this section, we show the impact of our models on PASCAL VOC. We first compare our method with the related state-of-the-art, then we carry out ablation study in order to understand the behavior of different components of our proposed sparse context transformer. We also show some qualitative results that highlight the impact of our models. In all these experiments, we take the average performance across 10 random runs.
	\subsubsection{Comparison}
	Table~\ref{tab:compareVOC} shows a comparison of  our method against the related state-of-the-art works which mostly report results with multiple random runs. Our proposed sparse context transformer obtains high accuracy on almost all the splits with different shots. Specifically, at extremely low-shot settings (i.e. 1-shot), our method achieves state-of-the-art performance on all the splits. These results demonstrate the ability of our proposed sparse context transformer to effectively combine contextual information of the target domain in order to overcome object class confusion in the few-shot detection scenarios.
	\begin{table}[htbp]
	\renewcommand{\arraystretch}{1.5}
	\centering
	\resizebox{0.5\textwidth}{!}{
		\begin{tabular}{c|c|c|c|c}
			\hline\hline\noalign{\smallskip}
			Method & context (w/ multi-scale feature) & Sparse relations & S & T \\
			\hline
			Baseline &  \XSolidBrush & \XSolidBrush & 68.1 & 26.0 \\
			\hline
			\multirow{3}{*}{Ours} & \Checkmark & \XSolidBrush & 69.7 &  30.1 \\
			&\XSolidBrush &\Checkmark & 69.7 & \textbf{32.4}\\ 
			&\Checkmark &\Checkmark & \textbf{69.9} &32.1 \\ 
			\hline\hline\noalign{\smallskip}
		\end{tabular}
	}\vspace{0.5cm}
	\caption{Ablation studies on the effectiveness of various components in our proposed sparse context transformer. The mAP with IoU threshold 0.5 (AP50) is reported. S and T stand for source and target domain categories respectively.}
	\label{AblationStudiesA}
\end{table}

\begin{table}[htbp]
	\renewcommand{\arraystretch}{1.5}
	\centering
	\resizebox{0.5\textwidth}{!}{
		\begin{tabular}{c|c|c|c}
			\hline\hline\noalign{\smallskip}
			& GAP & Attention Focus  & T \\
			\hline
			\multirow{3}{*}{\ \ \ \ \ Multi-scale Feature Maps \ \ \ \ \ } &\XSolidBrush 
			&\XSolidBrush & 31.0 \\
			& \Checkmark & \XSolidBrush  & 32.2 \\
			&\XSolidBrush &\Checkmark & \textbf{32.4} \\ 
			\hline\hline\noalign{\smallskip}	
		\end{tabular}
	}\vspace{0.5cm}
	\caption{The impact of our proposed attention focus layer. The experiments are conducted on the VOC 2007 test set of the PASCAL VOC dataset with novel split2 and AP50 on 1-shot task. Here T stands for target and GAP for global average pooling layer.}
	\label{AblationStudiesB}
\end{table}	

	\subsubsection{Ablation}
	In this ablation study, all the models are trained on the most difficult 1-shot scene set (in the novel set~2 of PASCAL VOC), and then evaluated on the PASCAL VOC 2007 test set. In these results, we again take the average performance across 10 random runs.\\
	
	\noindent\textbf{Impact of Context \& Sparsity}. Table~\ref{AblationStudiesA} (row 2 vs 1) shows the effectiveness of our proposed fusion approach of {\it contextual fields} from different sources. These results show that at extremely low-shot settings, the fusion of contextual fields from different sources improves the recognition performance of new categories in the target domain.  Table~\ref{AblationStudiesA} (row 3 vs 1) also demonstrates the positive impact of {\it sparsity}, i.e., constructing   sparse relationships between contextual fields and prior boxes. Indeed, at extremely low-shot settings, sparsity improves the accuracy of new object categories in the target domain. Besides, the combination of context and sparsity, at extremely low shot settings, not only improves the detector ability to recognize source domain classes but also target domain (new) ones. \\
		
	\begin{figure}[tbp]
	
	\centering 
	\subfigbottomskip=0.5pt
	\subfigure{\includegraphics[width=0.19\linewidth]{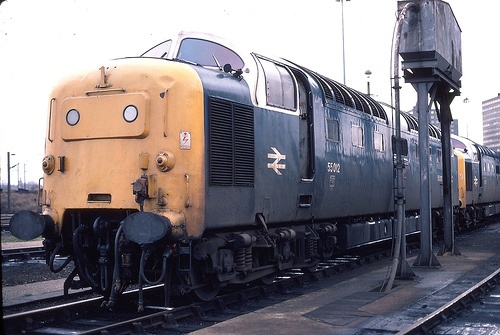}}
	\subfigure{\includegraphics[width=0.19\linewidth]{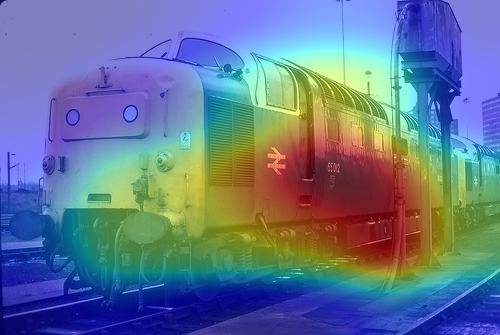}}
	\subfigure{\includegraphics[width=0.19\linewidth]{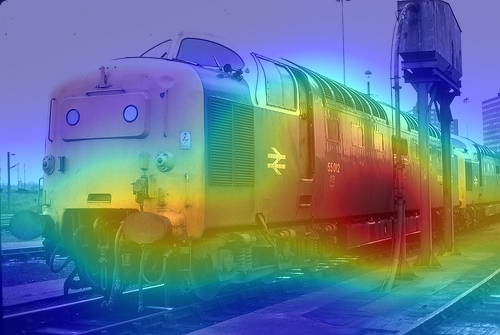}}
	\subfigure{\includegraphics[width=0.19\linewidth]{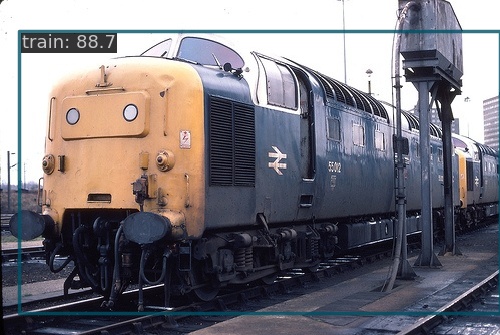}}
	\subfigure{\includegraphics[width=0.19\linewidth]{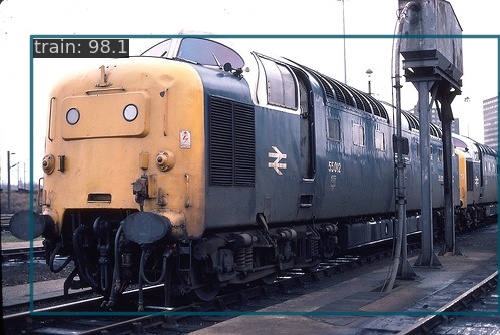}}
	\\
	\subfigure{\includegraphics[width=0.19\linewidth]{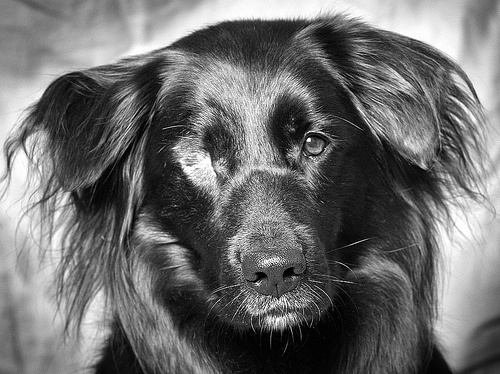}}
	\subfigure{\includegraphics[width=0.19\linewidth]{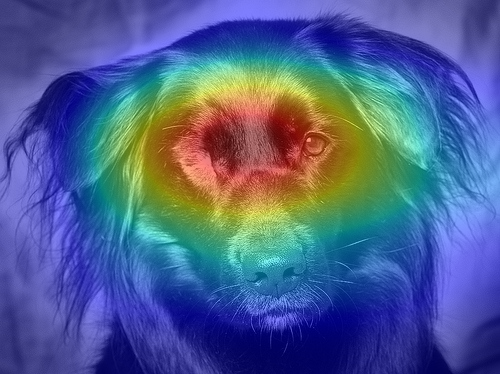}}
	\subfigure{\includegraphics[width=0.19\linewidth]{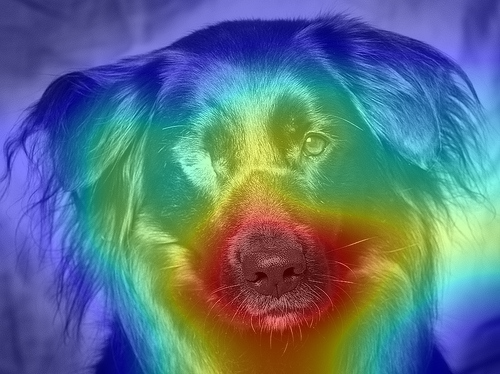}}
	\subfigure{\includegraphics[width=0.19\linewidth]{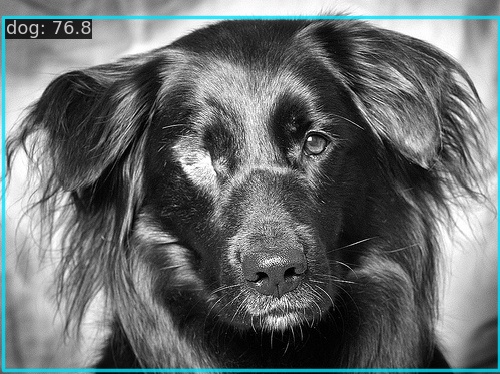}}
	\subfigure{\includegraphics[width=0.19\linewidth]{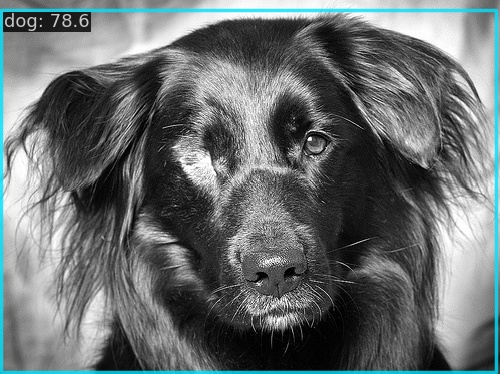}}
	\\
	\subfigure{\includegraphics[width=0.19\linewidth]{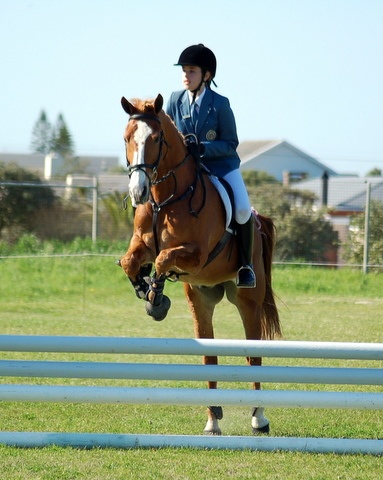}}
	\subfigure{\includegraphics[width=0.19\linewidth]{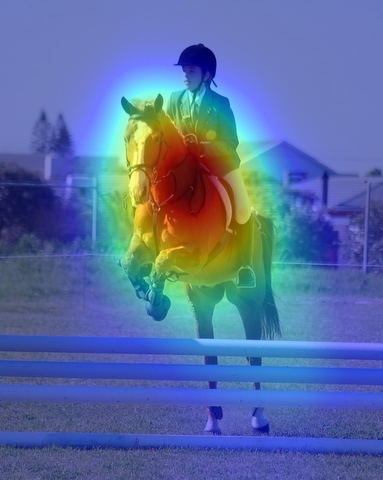}}
	\subfigure{\includegraphics[width=0.19\linewidth]{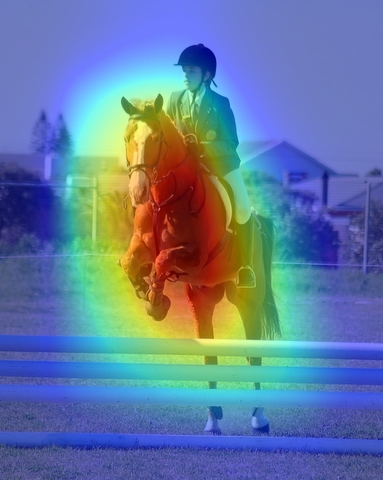}}
	\subfigure{\includegraphics[width=0.19\linewidth]{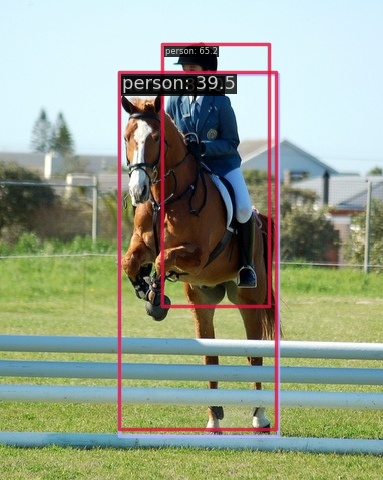}}
	\subfigure{\includegraphics[width=0.19\linewidth]{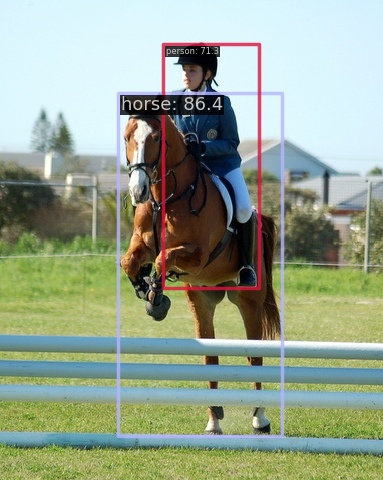}}
	\\
	\subfigure{\includegraphics[width=0.19\linewidth]{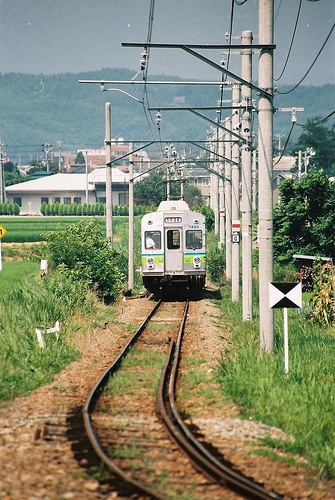}}
	\subfigure{\includegraphics[width=0.19\linewidth]{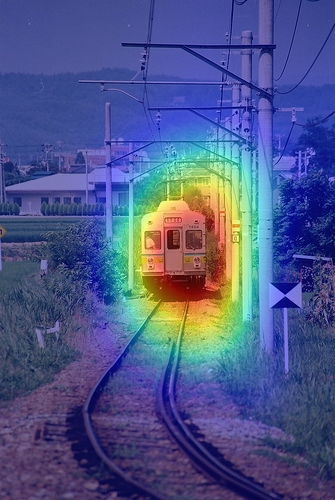}}
	\subfigure{\includegraphics[width=0.19\linewidth]{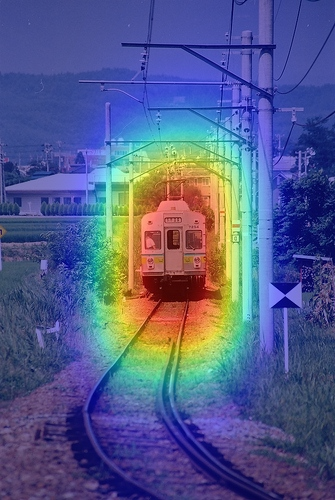}}
	\subfigure{\includegraphics[width=0.19\linewidth]{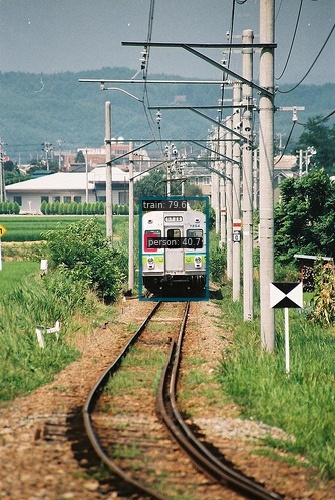}}
	\subfigure{\includegraphics[width=0.19\linewidth]{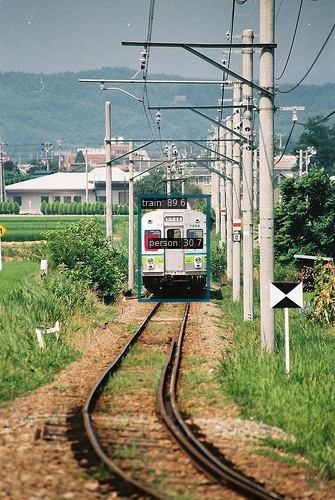}}
	\\
	\subfigure[\textbf{Input}]{\includegraphics[width=0.19\linewidth]{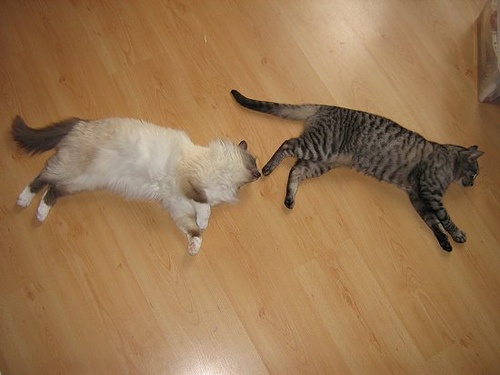}}
	\subfigure[\textbf{W/o SCT}]{\includegraphics[width=0.19\linewidth]{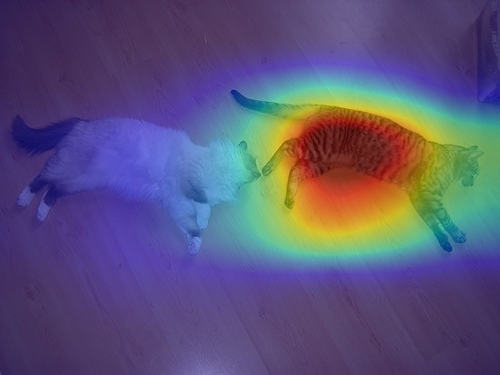}}
	\subfigure[\textbf{W/ SCT}]{\includegraphics[width=0.19\linewidth]{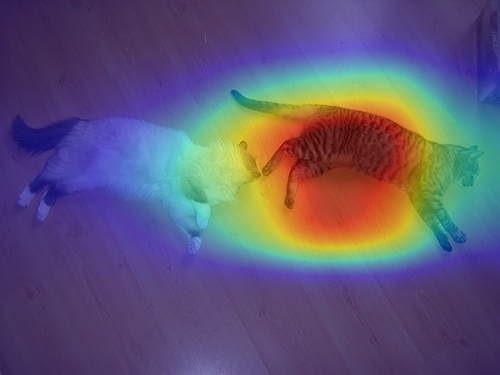}}
	\subfigure[\textbf{W/o SCT}]{\includegraphics[width=0.19\linewidth]{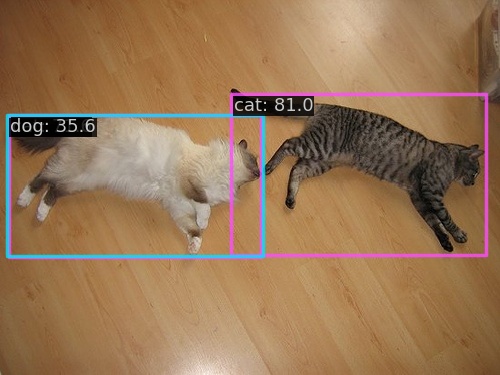}}
	\subfigure[\textbf{W/ SCT}]{\includegraphics[width=0.19\linewidth]{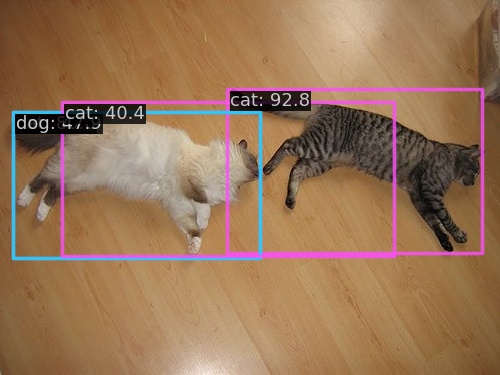}}
	\caption{Visualization of the results with (w/) and without (w/o)  Sparse Context Transformer (SCT) on the PASCAL VOC dataset.  Different colored bounding boxes represent different categories.}
	\label{qualitative}
\end{figure}
		
	\noindent	\textbf{Impact of Attention Focus}. Table~\ref{AblationStudiesB} (line 3 vs 1+2) shows the effectiveness of our designed attention focus layer. From these results, we observe a clear positive impact  of our proposed attention focus layer on target domain (new) classes compared to when no attention is considered.
	\begin{table*}[htbp]
		\centering
		\resizebox{0.9\textwidth}{!}{
			\begin{tabular}{c|cccccc|cccccc|}
				\hline\hline\noalign{\smallskip}	
				\multirow{2}{*}{Method/shot} & \multicolumn{6}{c|}{10-shot}  &\multicolumn{6}{c|}{30-shot} \\
				& AP & $ AP_{50} $ & $ AP_{75} $ & $ AP_{S} $ &  $ AP_{M} $ & $ AP_{L} $ & AP & $ AP_{50} $ & $ AP_{75} $ & $ AP_{S} $ &  $ AP_{M} $ & $ AP_{L} $\\
				\hline\hline\noalign{\smallskip}
				LSTD\cite{DBLP:journals/corr/abs-1803-01529}  & 3.2 & 8.1 & 2.1 & 0.9 & 2.0 & 6.5 & 7.8 & 10.4 & 10.4 & 1.1 & 5.6 & \textcolor{blue}{19.6} \\
				Meta YOLO \cite{Kang_Liu_Wang_Yu_Feng_Darrell_2019} & 5.6 & 12.3 & 4.6 & 0.9 & 3.5 & 10.5 & 10.1 & 14.3 & \textcolor{red}{14.4} & 1.5 & \textcolor{blue}{8.4} & \textcolor{red}{28.2} \\
				Baseline\cite{Yang_Wang_Chen_Liu_Qiao_2020} & \textcolor{blue}{7.7} & \textcolor{blue}{13.5} & \textcolor{blue}{7.5}& \textcolor{red}{2.3} & \textcolor{red}{5.4} & \textcolor{blue}{12.2} &\textcolor{blue}{11.0} & \textcolor{blue}{19.5} & 10.8 &\textcolor{blue}{2.1} & 8.3 & 18.6  \\
				SparseCT(ours) & \textcolor{red}{7.9} & \textcolor{red}{14.3} & \textcolor{red}{7.7} & \textcolor{blue}{1.9} & \textcolor{blue}{4.5} & \textcolor{red}{12.4} &\textcolor{red}{11.2} & \textcolor{red}{20.2} & \textcolor{blue}{11.3} & \textcolor{red}{2.2} & \textcolor{red}{8.5} & 19.3  \\
				\hline
			\end{tabular}
		}\vspace{0.5cm}
		\caption{Few-Shot detection performance on COCO minival of MS COCO dataset. We report the mean Averaged Precision on the 20 novel classes of COCO. \textcolor{red}{red} and \textcolor{blue}{blue} indicate the best and second best results.}
		\label{tab:compareCOCO}
	\end{table*}
	\begin{figure}[tbp]
		\subfigbottomskip=0.5pt
		\subfigure{\includegraphics[width=0.32\linewidth]{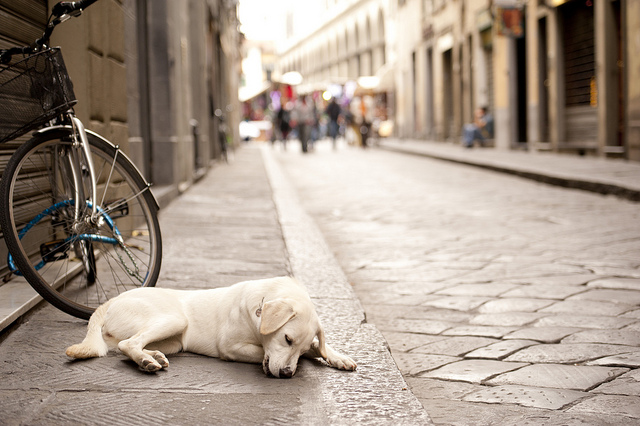}}
		\subfigure{\includegraphics[width=0.32\linewidth]{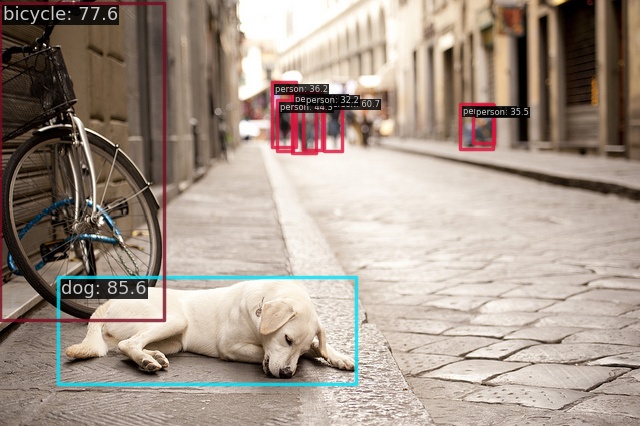}}
		\subfigure{\includegraphics[width=0.32\linewidth]{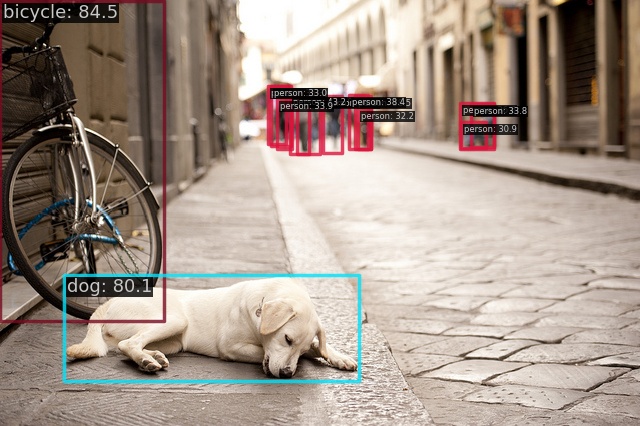}}
		\\
		\subfigure{\includegraphics[width=0.32\linewidth]{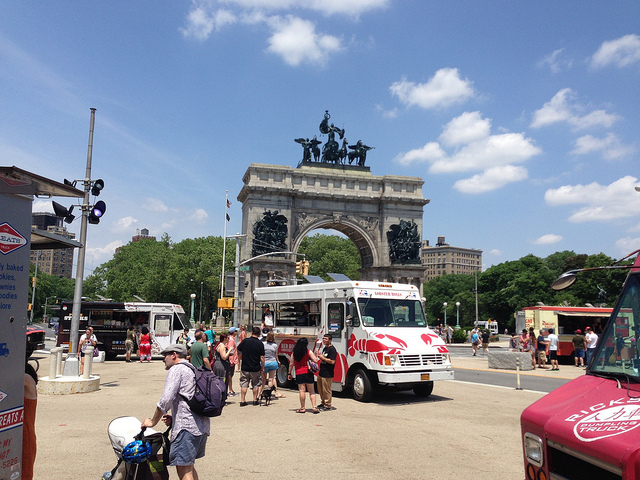}}
		\subfigure{\includegraphics[width=0.32\linewidth]{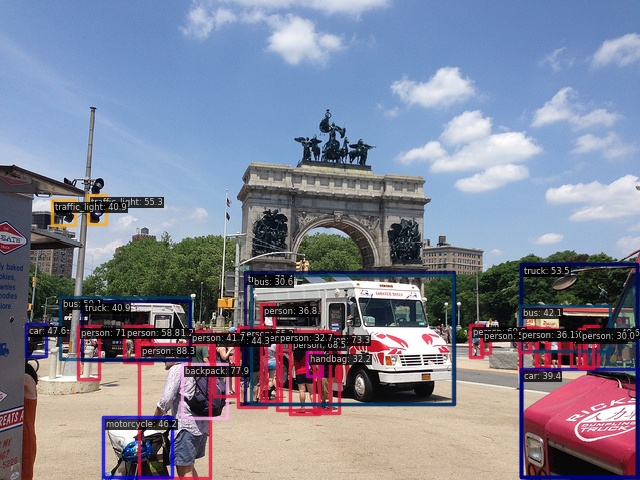}}
		\subfigure{\includegraphics[width=0.32\linewidth]{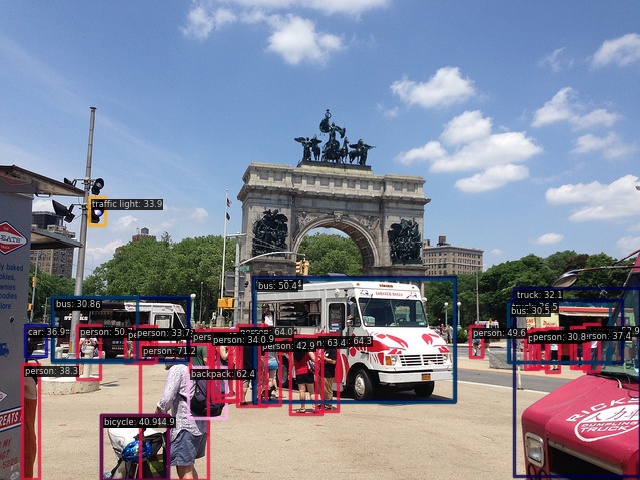}}
		\\ 
		\subfigure{\includegraphics[width=0.32\linewidth]{COCO/COCO_val2014_000000002988.jpg}}
		\subfigure{\includegraphics[width=0.32\linewidth]{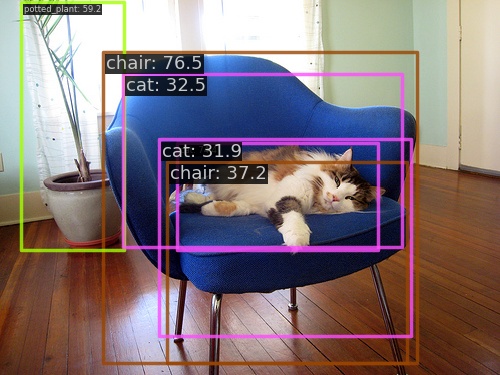}}
		\subfigure{\includegraphics[width=0.32\linewidth]{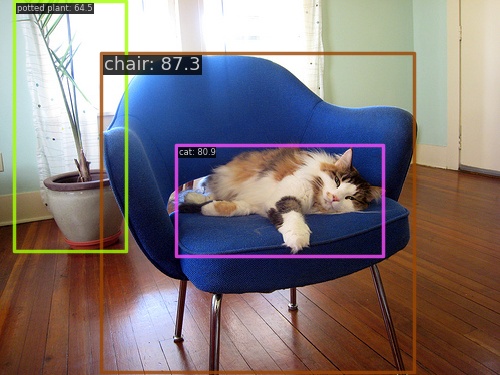}}
		\\
		\subfigure{\includegraphics[width=0.32\linewidth]{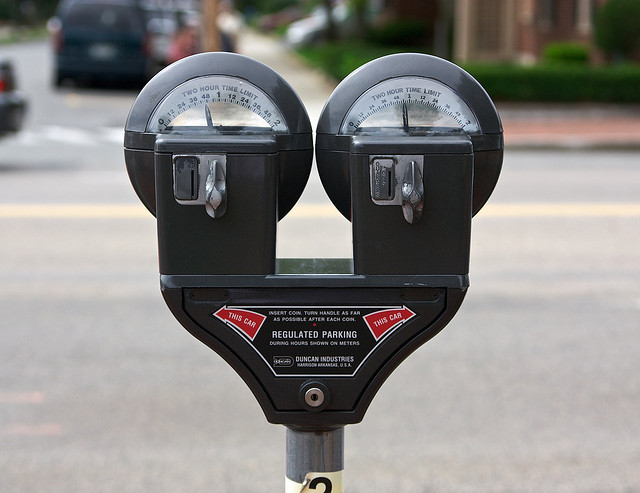}}
		\subfigure{\includegraphics[width=0.32\linewidth]{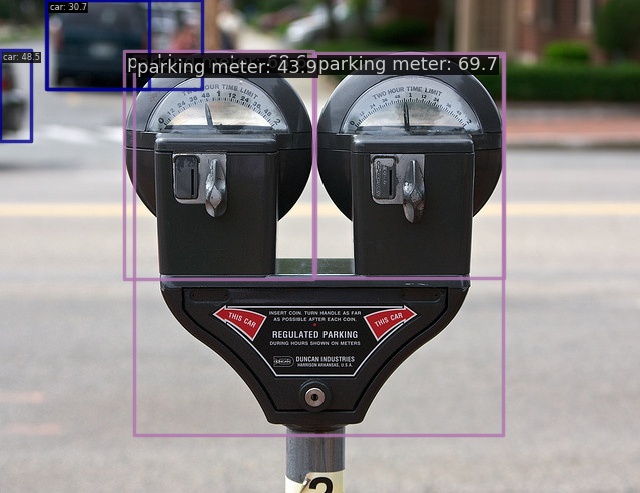}}
		\subfigure{\includegraphics[width=0.32\linewidth]{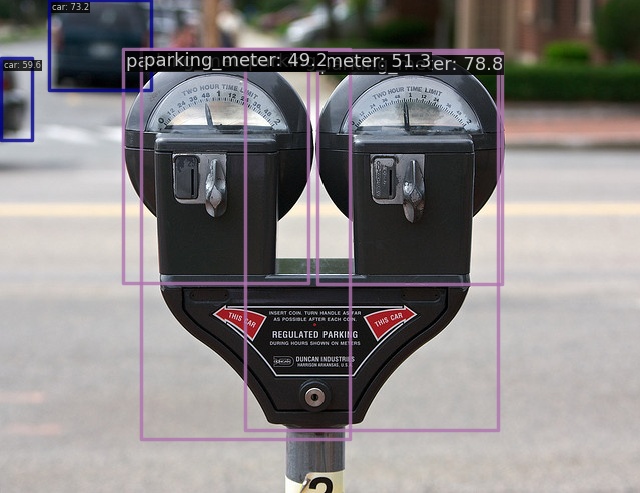}}
		\\
		\subfigure[\textbf{Input}]{\includegraphics[width=0.32\linewidth]{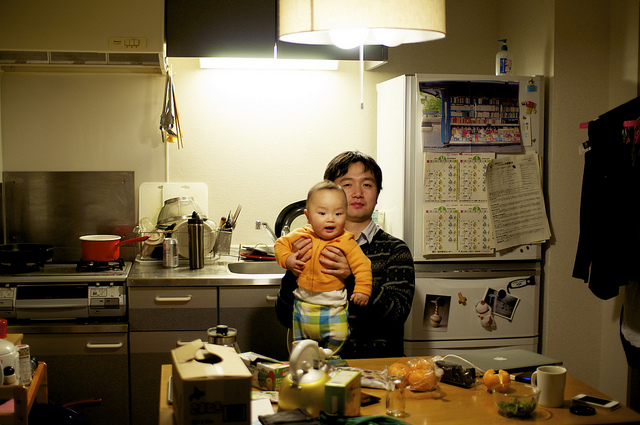}}
		\subfigure[\textbf{Baseline}]{\includegraphics[width=0.32\linewidth]{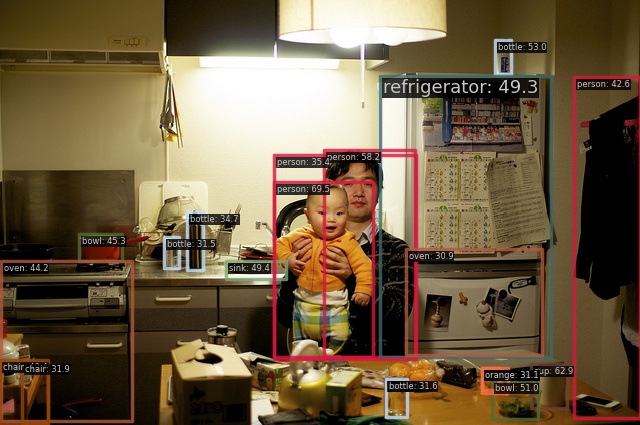}}
		\subfigure[\textbf{Ours}]{\includegraphics[width=0.32\linewidth]{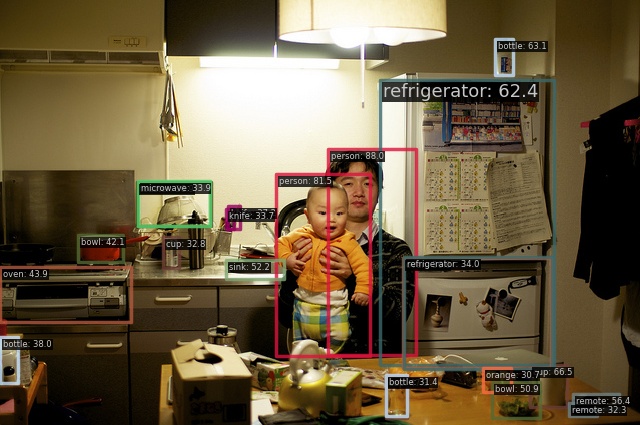}}
		\caption{This figure shows comparison results on the MS COCO dataset. Baseline represents the work of  Wang et al \cite{Yang_Wang_Chen_Liu_Qiao_2020}. Different colored bounding boxes represent different categories.}
		\label{qualitativeOnCOCO}
	\end{figure}
	\subsubsection{Qualitative Performance}	
	In order to further understand the impact  of our sparse context transformer module, we visualize the results with (w/) and without (w/o) SCT. As shown in Fig.~\ref{qualitative}, the activation maps in the second and third columns show that SCT improves attention to objects in images which eventually leads to better detection performance as shown in the fourth and fifth columns.  Furthermore, by comparing visualizations in the third row, we observe that SCT mitigates object confusion in few-shot detection.
	\subsection{Results on MS COCO}
	Finally, we provide extra performance on the 10/30-shot setups using the MS COCO benchmark, and we report the average performance using the standard COCO metrics over 10 runs  with random shows. Performance on novel classes are shown in Table~\ref{tab:compareCOCO}. Compared to the baseline, our proposed method improves performance in the 10/30 shot tasks, and is also competitive compared to the related works. Qualitative results are also shown in Fig.~\ref{qualitativeOnCOCO}; compared to the baseline in the related work~\cite{Yang_Wang_Chen_Liu_Qiao_2020}, our proposed method shows a noticeable improvement when handling confusion between detected object categories.

	\section{Conclusion}
	\label{sec:typestyle}
	In this paper, we propose a novel sparse context transformer that effectively explores useful contextual information for few shot object detection. The strength of the proposed model resides in its ability to learn sparse relevant relationships while discarding irrelevant ones. Extensive experiments conducted on two challenging standard benchmarks (namely PASCAL VOC and MS COCO) show the effectiveness of each component of our sparse context transformer and its outperformance with respect to the related work. \\ As a future work, we are currently investigating the extension of this model using other neural architectures, evaluation benchmarks and applications.

\end{document}